\documentclass{new_tlp}
\usepackage{mathptmx}

\usepackage{times}  
\usepackage{helvet}  
\usepackage{courier}  
\usepackage{url}  
\usepackage{graphicx}  
\usepackage{multicol}
\usepackage{multirow}
\usepackage{xspace}

\def\lar{\leftarrow}
\def\ba{\begin{array}}
	\def\ea{\end{array}}
\def\beq{\begin{equation}}
\def\eeq#1{\label{#1}\end{equation}}
\def\beqq{\begin{equation*}}
\def\eeqq{\end{equation*}}
\def\no{\ii{not}}
\def\ii#1{\hbox{\it #1\/}}

\def\lar{\leftarrow}
\def\ba{\begin{array}}
	\def\ea{\end{array}}
\def\beq{\begin{equation}}
\def\eeq#1{\label{#1}\end{equation}}
\def\beqq{\begin{equation*}}
\def\eeqq{\end{equation*}}
\def\no{\ii{not}}
\def\ii#1{\hbox{\it #1\/}}

\def\seq#1{\left\langle #1 \right\rangle}

\newcommand{\amp}[1]{\ensuremath{{\textsl{{\&}}}\!\,\mathit{#1}}}
\newcommand{\ext}[3]{\ensuremath{\amp{#1}[#2](#3)}}

\def\clingo{{\sc Clingo}}
\def\hcpasp{{\sc Hcp-Asp}}

\title[Theory and Practice of Logic Programming]
        {Human Robot Collaborative Assembly Planning: \\
        An Answer Set Programming Approach}

\author[Rizwan, Patoglu, Erdem]{
        Momina Rizwan\footnote{Contact Author}, Volkan Patoglu, and Esra Erdem\\
        Faculty of Engineering and Natural Sciences, Sabanc{\i} University, Istanbul, Turkey \\
        \{momina,volkan.patoglu,esra.erdem\}@sabanciuniv.edu}

\begin{document}

\label{firstpage}

\maketitle

\begin{abstract}
For planning an assembly of a product from a given set of parts, robots necessitate certain cognitive skills: high-level planning is needed to decide the order of actuation actions, while geometric reasoning is needed to check the feasibility of these actions. For collaborative assembly tasks with humans, robots require further cognitive capabilities, such as commonsense reasoning, sensing, and communication skills, not only to cope with the uncertainty caused by incomplete knowledge about the humans’ behaviors but also to ensure safer collaborations. We propose a novel method for collaborative assembly planning under uncertainty, that utilizes hybrid conditional planning extended with commonsense reasoning and a rich set of communication actions for collaborative tasks. Our method is based on answer set programming. We show the applicability of our approach in a real-world assembly domain, where a bi-manual Baxter robot collaborates with a human teammate to assemble furniture. This manuscript is under consideration for acceptance in TPLP.
\end{abstract}



\section{Introduction}

As high scale industries move towards customized products, robotic assembly tasks become not only physically, but also mentally challenging. For this reason, drastic changes have been taking place for industrial robotics over the past few years. While working areas of humans and robots were strictly separated in the past, nowadays  collaboration among robots and human operators are necessitated such that flexible assembly systems can benefit from both the precision of robots and the adaptability of humans. Human-robot interactions need to be safe and socially appropriate to lead to improved performance and team satisfaction. However, the involvement of humans in the robot workplace poses many challenges due to uncertainty about the actions, behaviors and intentions of
humans.

Collaborative assembly planning to produce customized products necessitates robots to possess certain cognitive abilities. For instance, for assembly planning, high-level task planning is required to decide for the order of actuation actions (e.g., picking, holding, joining, placing), while sensing is required to resolve uncertainty due to incomplete knowledge about the world (e.g., to check for existence of proper connections). Meanwhile, geometric reasoning is required to ensure the feasibility of both actuation and sensing actions (e.g., checking whether there exists a collision-free path to perform a pick action). In addition, for collaborations with humans, robots need to be furnished with further cognitive capabilities, including commonsense reasoning (e.g., knowing that humans cannot carry heavy parts), sensing to resolve uncertainty about human actions (e.g., checking whether the human is holding a part to be assembled), and communication skills to resolve uncertainty about human intentions and to ensure safe and socially acceptable interactions. These communication skills involve greetings, asking/offering help, confirming intentions, requesting actions, warnings, and providing explanations.  Endowing robots with such a variety of cognitive capabilities make collaborative assembly planning even more challenging.

We propose a novel method for collaborative assembly planning, utilizing hybrid conditional planning (\hcpasp)~\cite{yalcinerNPE17} based on answer set programming (ASP)~\cite{BrewkaEL16}.

\hcpasp\ enables offline planning of actuation and sensing actions starting from an initial state to reach a goal state, in the presence of incomplete knowledge and partial observability, by considering all possible contingencies, and by considering feasibility of actions. The computed plans can be viewed as trees of actuation actions, whose effects are deterministic, and sensing actions, whose effects are non-deterministic. Each branch of the tree from the root to a leaf represents a possible execution of actuation and sensing actions to reach a goal state from the given initial state.

The novelties of our approach to collaborative assembly planning can be summarized as follows, along with our contributions:

\smallskip\noindent {\em Hybrid actuation and sensing actions.}
To solve collaborative assembly planning problems using \hcpasp, we model relevant actuation and sensing actions in ASP, and, in particular, illustrate how continuous geometric feasibility checks (e.g., collision-free reachability checks) can be embedded directly into logical descriptions of these actions by means of hard constraints.

\smallskip\noindent {\em Communication actions.}
Since collaborative assembly planning necessitates more interactive collaborations between a robot and a human, we extend \hcpasp\ to include communication actions. These actions are different from actuation and sensing actions from several perspectives, and thus we present a novel method for modeling them.
\vspace{-1mm}
\begin{itemize}
\item Effects.
These actions are different from actuation and sensing actions from the perspective of their effects:  some of the communication actions have deterministic effects, while some have nondeterministic effects. For instance, requesting a cooperative human teammate to perform some action, initiating/ending conversations, and providing explanations have deterministic effects. On the other hand, confirming some actions, asking for help, and offering some help necessitate some answers/feedback from human, and thus have nondeterministic effects.
Due to these differences, we identify five types of communication actions relevant for collaborative assembly planning, and introduce a method for modeling the effects of each type of communication action.

\item Preconditions.
These actions are also different from actuation and sensing actions from the perspective of their preconditions: while the preconditions of actuation and sensing actions are concerned about their executability, the preconditions of communication actions involve commonsense knowledge for a more natural human-robot interaction (e.g., not asking for help if the human is busy), as well as safety concerns (e.g., not asking for help in attaching a part, if that part is dangerous for a human to touch). We identify relevant commonsense knowledge and safety concerns, and model the preconditions of communications accordingly to compute human-aware plans.

\item Feasibility checks.
In actuation and sensing actions, feasibility checks are added as hard constraints, as the robot is  not capable of performing such actions physically otherwise. However, in collaborative assembly problem, the robot can resolve its inability to perform an action by asking for help from the human teammate when the robot fails to perform a task. For that reason, feasibility checks are embedded in communication actions differently, by utilizing weak constraints.
\end{itemize}

\noindent {\em Empirical evaluation.}
To investigate the usefulness and scalability of our approach, we perform experiments over a furniture assembly domain that involves collaborations between a robot and a human. Considering different types of human-robot interaction, we vary the number of unsafe parts, parts that are reachable by the human teammate only, and parts that are reachable by the robot only.

\smallskip\noindent {\em A real-world application.}
We illustrate applications of our method over a collaborative furniture assembly planning domain, where a bi-manual Baxter robot collaborates with a human teammate to assemble a coffee table.

\section{Answer Set Programming}

We use Answer Set Programming (ASP)~\cite{BrewkaEL16}---a logic programming paradigm based on answer sets---for hybrid conditional planning as described in~\cite{yalcinerNPE17}. Let us go over some special constructs of ASP used in our study, before we describe its use.

We consider rules of the form
$$
\ii{Head} \lar A_1, \dots, A_m, \no\ B_{m+1}, \dots, \no\ B_n
$$
where $n \geq m \geq 0$, \ii{Head} is a literal (a propositional atom p or its negation $\neg p$) or $\bot$, and each $A_i$ is an atom or an external atom~\cite{hex2005}. A rule is called a \textit{fact} if $m=n=0$, and a \textit{constraint} if \ii{Head} is
$\bot$. A set of rules is called a \textit{program}.

An external atom is an expression of the form $\ext{g}{y_1,\dots,y_k}{x_1,\dots,x_l}$ where $y_1,\dots,y_k$ and $x_1,\dots,x_l$ are two lists of terms (called input and output lists, respectively), and $\&g$ is an external predicate name. Intuitively, an external atom provides a way for deciding the truth value of an output tuple depending on the extension of a set of input predicates. External atoms allow us to embed results of external computations into ASP programs.
For instance, the following rules express that, at any step $t$ of the plan, a robot cannot place an object $o$ at location $(x_1,y_1)$ if there is no collision-free trajectory between them:
$$\ba l
\lar \ii{place}(a,x_1,y_1,t), \ii{holding}(a,o,t),  \no\ \ext{\ii{collision\_free}}{a,x_1,y_1}{}
\ea
$$
The external atom $\ext{\ii{collision\_free}}{a,x_1,y_1}{}$ takes $a$, $x_1$, $y_1$ as inputs to an external function implemented in Python. This external function calls a motion planner (e.g., the RRT* motion planner~\cite{rrt-star2011} from OMPL~\cite{sucan2012the-open-motion-planning-library} library) to check the existence of a collision free trajectory for the arm $a$ to reach $(x_1,y_1)$. Then, it returns the result of the computation (i.e., True or False) as a precondition.

ASP provides special constructs to represent a variety of knowledge. For instance, it is possible to express nondeterministic choice in ASP using ``choice expressions'' with ``cardinality constraints.'' Choice expressions help us to model occurrences and non-occurrences of actions. For instance, the following ASP rule expresses that the action of sensing the location of an object can occur any time:
$$
\ba l
\{\ii{sense}(\ii{at}(o),t)\} .
\ea
$$

Choice expressions with cardinality constraints help us to model nondeterministic effects of sensing actions. For instance, the following ASP rule describes that if sensing is applied to check the location of an object $o$ (i.e., $\ii{sense}(at(o),t) $), then we know that the object $o$ is at one of the possible locations $l$:
$$
\ba l
1\{\ii{at}(o,l,t+1): loc(l) \}1 \lar \ii{sense}(\ii{at}(o),t) .
\ea
$$
Here, the location $l$ is nondeterministically chosen by the ASP solver.

Also, it is possible to express ``unknowns'' using ``cardinality expressions''. For instance, the following rule expresses that if the location of object $o$ is not known (i.e., $\{ \ii{at}(o,l,t) : loc(l) \}0$), then it definitely can not be at a robot’s hand $m$:
$$
\ba l
\neg \ii{at}(o,m,t) \lar \{ \ii{at}(o,l,t) : loc(l) \}0
\ea
$$

In addition to choice rules and cardinality expressions, we also utilize ``weak constraints'' to express preferences over occurrences of types of actions in a plan. For instance, the following weak constraint minimizes the number of sensing actions:
$$
:\sim \ii{senseAct}(t)\  [2@2,t].
$$

\section{Hybrid Conditional Planning}

Conditional planning enables planning from an initial state to a goal state in the presence of incomplete knowledge and partial observability~\cite{Warren1976,PeotS92,PryorC96} by considering all possible contingencies. Thus the plans (called conditional plans) are trees of actuation actions, whose effects are deterministic, and sensing actions, whose effects are non-deterministic, where each branch of the tree from the root to a leaf represents a possible execution of actuation and sensing actions to reach a goal state from the given initial state.

A hybrid conditional planner allows us to ensure that there are no physical constraints while executing the computed hybrid conditional plan, by introducing external computation during planning phase to determine feasibility of each action. As a result, infeasible actions are removed from the plan to prevent failure of a branch.

\begin{figure}[b]
	\centering
	\begin{tabular}{cc}
    \resizebox{0.5\columnwidth}{!}{\rotatebox{0}{\includegraphics{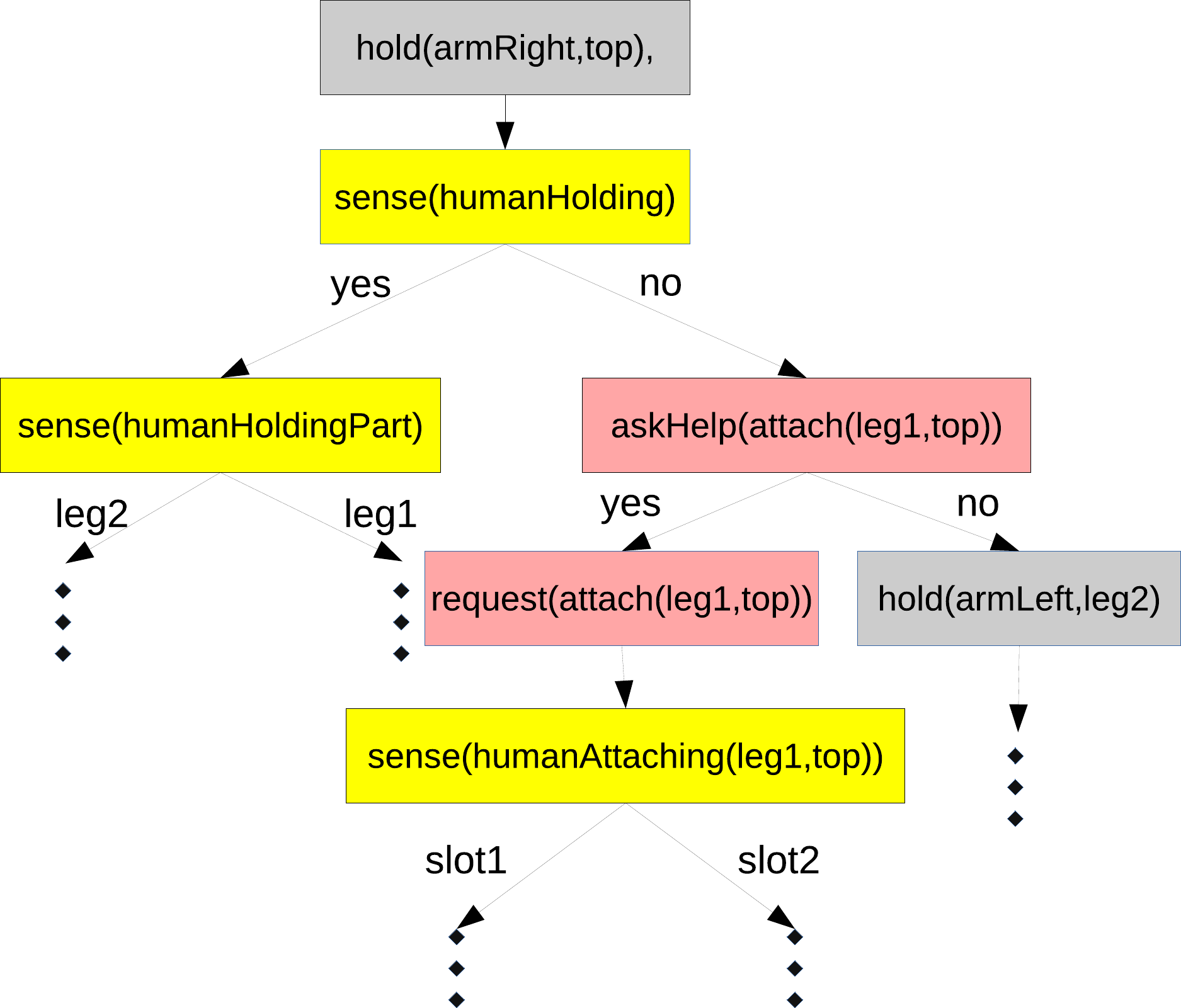}}} &
    \includegraphics[scale=0.2]{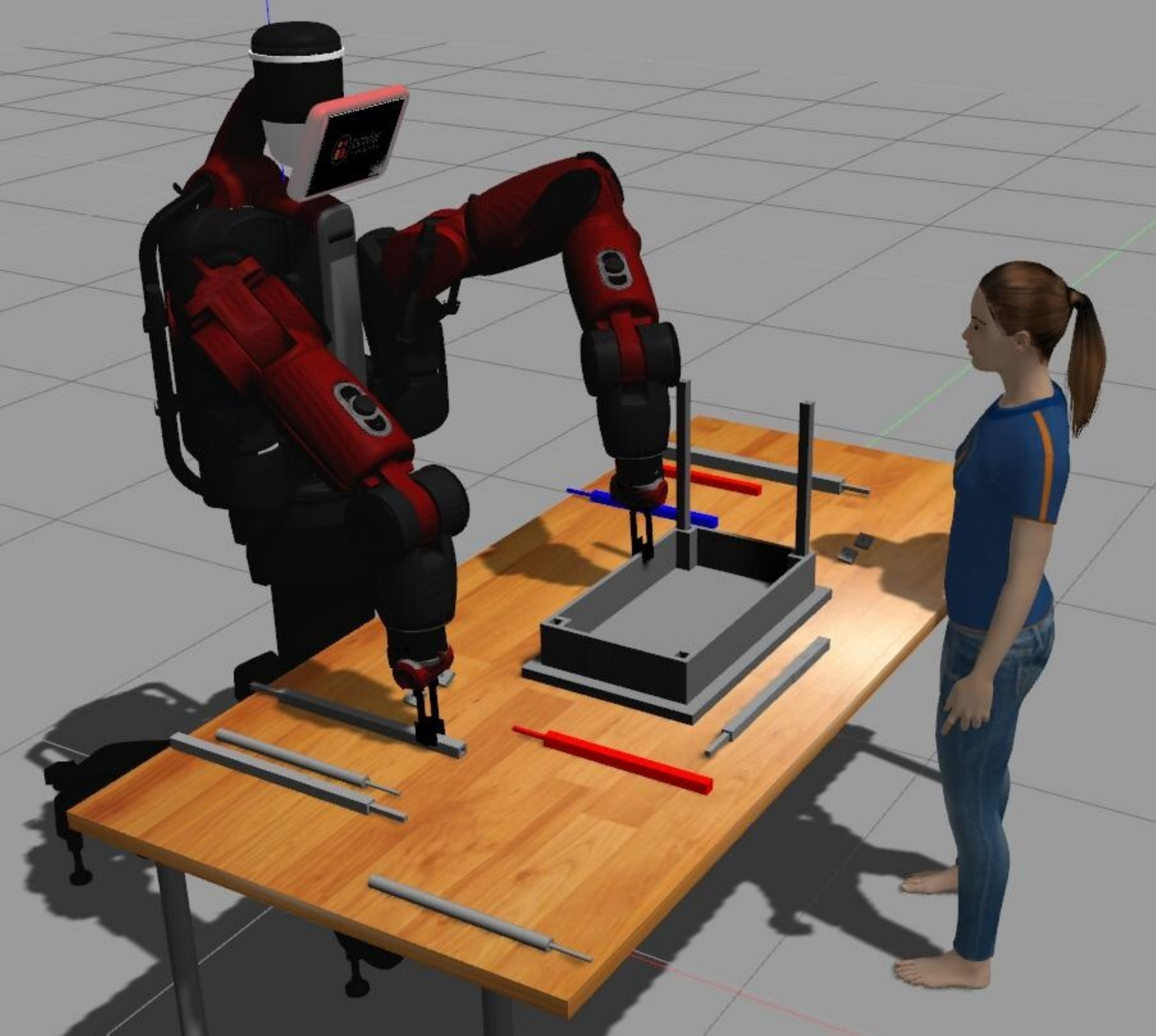} \\
    (a) & (b)
    \end{tabular}\vspace{-.5\baselineskip}
	\caption{(a) Some part of a hybrid conditional plan computed for (b) a human-robot collaborative assembly planning scenario.}	
	\label{fig:cplanTree2}
\end{figure}

A hybrid conditional plan can be identified as a labeled directed tree $(V,E)$ as in Figure \ref{fig:cplanTree2} where every branch represents a possible executable plan. The set $V = V_a \cup V_s$ of vertices denote actions in the conditional plan consisting of two types of vertices. The vertices in $V_a$ represent hybrid actuation actions (e.g., the robot's manipulation actions to hold an object integrated with reachability checks) are highlighted as gray in Figure \ref{fig:cplanTree2}. Whereas the vertices in $V_s$  represent sensing actions (e.g., sensing the shape or color of an object) highlighted as yellow in Figure \ref{fig:cplanTree2}. The branching occurs when there is a sensing action with non-deterministic outcome, so every vertex in $V_{s}$ has at least two outgoing edges, while each vertex in $V_a$ has a single outgoing edge based on the assumption that the actuation actions are deterministic. Each sensing action may lead to different outcomes/observations.

The set of edges $E$ represents the order of actions in the directed graph. Let us denote by $E_s$ the set of outgoing edges from vertices in $V_s$. Then a labeling function maps every edge $(x,y)$ in $E_s$ by a possible outcome of the sensing action characterized by~$x$.

In this study, we use the hybrid conditional planner HCP-ASP~\cite{yalcinerNPE17}, based on a parallel algorithm that calls the ASP solver \clingo\ to compute the branches. The actuation actions and sensing actions are represented in ASP, and the feasibility checks are embedded into these action descriptions by external atoms, as suggested by~\citeN{yalcinerNPE17}.

\section{Representing Assembly Planning in ASP: No Communications}

In an assembly domain, world states are described by fluents (i.e., atoms whose value change by time). Some of these fluents are fully observable (i.e., the robot knows their values) and their values are determined by actuation actions. Some of the fluents are partially observable (i.e., the robot may not know their values) and their values are determined by sensing actions.

For instance, in the table assembly domain, the fully observable fluent $\ii{attached}(p,p',c,t)$ represents that part $p$ is attached to part $p'$ at connection point $c$ at time step $t$. The fully observable fluent $\ii{holding}(m,p,t)$ represents that manipulator $m$ of the robot is holding part $p$ at time step $t$. The values of fully observable fluents are determined by the actuation actions:
\begin{itemize}
	\item $\ii{hold}(m,p,t)$ (hold the part $p$ with the manipulator $m$ at time step $t$),
	\item $\ii{attach}(m,p',c,t)$ (attach the part being held by the manipulator $m$ to the part $p'$ through connection point $c$ at time step $t$), and
	\item $\ii{unhold}(m,t)$ (un-hold the part being held by the manipulator $m$ at time step $t$).
\end{itemize}

The partially observable fluent $\ii{humanHolding}(t)$ describes that the human is holding something at time $t$, the partially observable fluent  $\ii{humanHoldingPart}(p,t)$ describes that the human is holding a part $p$ at time $t$, and the partially observable fluent  $\ii{humanAttaching}(p,p',c,t)$ describes that the human is attaching part $p$ to $p'$ at attach point $c$ at time step $t$. The values of these fluents are determined by the following sensing actions:
\begin{itemize}
	\item
	$\ii{sense}(humanHolding,t)$ (sense if the human is holding anything or not at time step~$t$),
	\item
	$\ii{sense}(humanHoldingWhichPart,t)$ (sense which part human is holding at time step~$t$),
	\item
	$\ii{sense}(humanUnholding(p),t)$ (sense if human is unholding part $p$ at time step~$t$), and
	\item
	$\ii{sense}(humanAttachingWhere(p,p'),t)$ (sense where the human is attaching the parts $p$ and $p'$ at time step~$t$).
\end{itemize}

The actuation actions and sensing actions are represented in ASP for hybrid conditional planning, as described in~\cite{yalcinerNPE17}.
For instance, consider the robot's action of holding the assembly part $p$ at time step $t-1$. As a deterministic effect of this action, the part $p$ will be in robot's hand at the next time step $t$:
$$
\ii{holding}(m,p,t) \lar \ii{hold}(m,p,t{-}1).
$$
Similarly, as a direct effect, $\ii{attach}$ action will join part $p$ in the robot's hand to a part $p'$ at the attach point $c$,
$$
\ii{attached}(p,p',c,t) \lar \ii{attach}(m,p',c,t{-}1), \ii{holding}(m,p,t{-}1).
$$
The preconditions of actuation actions are represented by constraints. For instance, a manipulator cannot hold a part $p$, if the manipulator is not free:
$$
\lar \ii{hold}(m,p,t), \no\ \ii{free}(m,t).
$$
A manipulator $m$ cannot attach a part $p'$ to a part $p$, if it is not already holding $p'$. In this case, we represent this precondition by projecting $\ii{attach}(m,p,c,t)$ to $\ii{attachPRT}(m,p,t)$:
$$
\ba l
\lar \ii{attachPRT}(m,p,t),\ \{\ii{holding}(m,p',t): \ii{parts}(p'), p \neq p'\}0.
\ea
$$

Sensing actions are represented by atoms of the form $\ii{sense}(f,t)$, where $f$ is a partially observed fluent. The nondeterministic effects of sensing actions are described using atoms of the form $\ii{sensed}(f'',t)$, where $f''$ denotes the relevant partially observed fluent, within choice rules.  A nondeterministic effect of robot observing whether the human is holding something can be formulated by the following choice rule:
$$
\{\ii{sensed}(\ii{humanHolding},t)\} \lar \ii{sense}(\ii{humanHolding},t-1).
$$
Suppose that the human can only hold one part at a time.  A nondeterministic effect of robot observing which part the human is holding, can be formulated by the following choice rule:
$$
1\{\ii{sensed}(\ii{humanHoldingPart}(p),t): \ii{parts}(p)\}1 \lar \ii{sense}(\ii{humanHoldingWhichPart},t-1).
$$
The preconditions of sensing actions are also described by constraints. For instance, the robot can observe which part the human is holding, if the robot has already sensed that the human is holding something:
$$
\lar \ii{sense}(\ii{humanHoldingWhichPart},t), \no\ \ii{humanHolding}(t)
$$
where $\ii{humanHolding}(t)$ is defined as follows:
$$
\ii{humanHolding}(t) \lar \ii{sensed}(\ii{humanHolding},t) .
$$

The feasibility checks are embedded in the descriptions of actuation actions and sensing actions, using external atoms. For instance, in the table assembly domain, the robot can hold a part if there exists a kinematic solution to reach the part with its manipulator. Such a reachability check can be embedded in the precondition of $\ii{hold}$ actions as follows:
$$
\lar \ii{hold}(m,p,t), \ii{loc}(p,r,t), \no\ \ext{\ii{reachable}}{m,r}{}.
$$
In these constraints, the reachability check is performed by the external atom $\ext{\ii{reachable}}{m,r}{}$, which calls a bi-directional RRT* motion planner~\cite{rrt-star2011} from OMPL~\cite{sucan2012the-open-motion-planning-library} library via a Python program to check for the collision-free forward kinematics solution to reach region $r$ with the manipulator $m$. Such an external atom returns true if there exists a collision-free trajectory to reach region $r$, and false otherwise.

With such a description of the table assembly domain, the robot can find a plan using the hybrid conditional planner~\hcpasp. The collaboration between the robot and the human teammate solely relies on the robot's sensing actions. We extend this method to include communication actions.

\section{Communication Actions for Collaborative Assembly Planning}

Communication actions are required to resolve the uncertainty caused due to the incomplete knowledge about the human intentions and desires. More importantly, communication is needed in a collaborative planning system to provide fluent and socially appropriate collaboration.
For this reason, in addition to actuation actions and sensing actions, we consider the following types of communication actions for collaborative table assembly domain:
\begin{itemize}
	\item [(i)] $\ii{confirmAttach}(p,p'),t)$ (confirming if human wants to attach $p$ to $p'$ at time step $t$)
	\item [(ii)] $\ii{askHelp}(p,p',t)$ (asking human help in attaching part $p$ to $p'$ at time step $t$)
	\item [(iii)] $\ii{offerHelp}(p,p',t)$ (offering help in attaching part $p$ to $p'$ at time step $t$)
	\item [(iv)] $\ii{requestToUnhold}(p,t)$ (requesting human to un-hold part $p$ at time step $t$)
	\item [(v)] $\ii{requestToAttach}(p,p',t)$ (requesting human to attach part $p$ to part $p'$ at time step $t$)
\end{itemize}
To describe the effects of these communication actions, we extend our list of partially observed fluents.

\smallskip\noindent{\em Effects of communication actions.} Communication actions are different from actuation and sensing actions, in that some of them are deterministic and some are nondeterministic. So we represent the direct effects of each communication action, depending on its type.

Requesting a collaborative human teammate to perform some action, initiating/ending conversations, and providing explanations have deterministic effects. Therefore, they are formalized as deterministic actions, like actuation actions. For instance, the effect of requesting a human teammate to attach a  part $p$ to another part $p'$ at time $t$ is represented as follows:
$$
\ii{requestedAttach}(p,p',t) \lar \ii{requestToAttach}(p,p',t-1).\quad (p\neq p')
$$

On the other hand, communication actions (e.g., asking for confirmation) that require some answers/feedback from humans are modeled as nondeterministic actions, like sensing actions. The nondeterministic communication actions serve as decision nodes in a hybrid conditional plan, similar to sensing actions. For instance, when the robot is unable to reach a part $p$, the robot asks the human teammate for some help in attaching a part $p'$ to part $p$. In return, the human responds affirmatively or negatively. The effect of asking for help in attaching part $p'$ to part $p$ is represented as follows:
$$
\ba l
1\{\ii{acceptToAttach}(p,p',t);\ \neg \ii{acceptToAttach}(p,p',t)\}1 \lar \\
\qquad \ii{askHelp}(p,p',t-1). \quad (p\neq p')
\ea
$$
Similarly, after the robot tries to confirm with the human as to whether she is planning to attach a part $p'$ to another part $p$, the human teammate may respond affirmatively or negatively:
$$
\ba l
1\{\ii{wantToAttach}(p,p',t+1);\ \neg \ii{wantToAttach}(p,p',t+1)\}1 \lar \\
\qquad \ii{confirmAttach}(p,p',t).\quad (p\neq p')
\ea
$$

\smallskip\noindent{\em Preconditions of communication actions: commonsense knowledge.}
All communication actions have relevant preconditions to ensure that they are executed when the appropriate conditions hold. For communication actions, most of the preconditions are due to commonsense knowledge. For instance, the robot can ask the human teammate for help in assembling a part $p$ to another $p'$ if the human is not already holding something:
$$
\ba l
\lar \ii{askHelp}(p,p',t), \ii{humanHolding}(t) \quad (p\neq p').
\ea
$$

If the human teammate is holding a part $p$,  the robot does not need to confirm that the human will be attaching $p$ to the part $p'$ that the robot is holding, if these two parts cannot be attached.
$$
\ba l
\lar \ii{confirmAttach}(p,p',t), \ii{class}(cl,p), \ii{class}(cl',p'), \no\ \ii{attachable}(cl,cl').
\ea
$$

\smallskip\noindent{\em Embedding feasibility checks.}
In actuation and sensing actions, feasibility checks are added as hard constraints as the robot is not physically capable of performing such actions otherwise. However, in collaborative assembly problem, the robot can resolve its inability to perform an action by asking for help from the human teammate when the robot fails to perform a task. To enable communication for such cases, we do not add a reachability check as a hard constraint, but include it as a weak constraint. We want to penalize such failures as much as possible. If such failures cannot be avoided, then they act as a precondition for the communication actions.

For instance, for reachability checks, we define failures as follows:
$$
\ba l
\ii{reachabilityFail}(m,p) \lar \ii{hold}(m,p,t),\ \ii{loc}(p,r,t), \no\ \ext{\ii{reachable}}{m,r}{}.
\ea
$$
and include the following weak constraint in the domain description:
$$
\ba l
:\sim \ii{reachabilityFail}(m,p).[2@1]
\ea
$$
This weak constraint penalizes a solution whenever a reachability check fails but still provides the best possible plan with a minimum number of reachability failures.
Then, the robot can only ask for help in attaching a part, if the task is infeasible for the robot (i.e., the robot cannot reach the part using any of its manipulators) and safe for the human teammate.
$$
\ba l
\lar \ii{askHelp}(p,p',t), loc(p,r,t),  \\
\qquad \no\ 2\{\ii{reachabilityFail}(m,p,t): \ii{manipulator}(m)\}2;\ \no\ \ii{unsafeRegion}(r).\\
\ea
$$

\smallskip\noindent{\em Safety.} Safety is an important concern for human-robot interactions. For instance, the robot should not ask the human teammate to attach a part $p$ (e.g., a wooden table leg with nails), which is dangerous for a human, to some other part $p'$. This can be expressed by the following constraint:
$$
\ba l
\lar \ii{askHelp}(p,p',t), \ii{type}(Dangerous,p).
\ea
$$

\section{Experimental Evaluations}

\noindent{\em Setup.}
In our experiments, we have used the HCP planner \hcpasp~\cite{yalcinerNPE17} for generating conditional plans, and RRT* motion planner~\cite{rrt-star2011} from OMPL~\cite{sucan2012the-open-motion-planning-library} for the reachability checks embedded into action descriptions. All experiments are performed on a Linux server with 12 2.4GHz Intel E5-2665 CPU cores and 64GB memory.

\begin{table}[t]
	\caption{Experimental results demonstrating the effect of increasing the number \#$U$ of unsafe parts on the size of the tree and the computation time (the time spent for planning and feasibility checks). For every tree, the number of nodes corresponding to each communication actions (i.e., Ask for help (K), Offer help (O), Confirm (C), and Request (R)) is reported as well.}
    \label{tab:cplanEE}
    \begin{minipage}{\textwidth}
		\begin{tabular}{ccccccccccccc}
			\hline
			Inst. & \#$U$ & $L$ & $D=A+S+C$ & \multicolumn{4}{c}{Communication actions} & $DN$ & $BF$ & $N$ & \multicolumn{2}{c}{Time (sec)} \\
			&  &  &  & K & O & C & R &  &  &  & Plan  &  Checks \\ \hline
			1 & 2 & 17 & 8+6+11 & 1 & 3 & 6 & 1 & 104 & 4 & 346 & 588 & 31 \\
			2 & 3 & 19 & 9+6+15 & 1 & 4 & 7 & 3 & 125 & 5 & 521 & 989 & 44 \\
			3 & 4 & 20 & 11+8+16 & 1 & 5 & 6 & 4 & 344 & 6 & 634 & 1339 & 46 \\
			4 & 5 & 24 & 12+7+17 & 1 & 6 & 6 & 4 & 432 & 7 & 777 & 3281 & 57 \\
			5 & 6 & 29 & 14+7+21 & 1 & 7 & 7 & 6 & 511 & 8 & 1123 & 5873 & 59 \\ \hline
		\end{tabular}
    \end{minipage}\vspace{-0.15\baselineskip}
\end{table}

\begin{table}[t]
	\caption{Experimental results demonstrating the effect of increasing the number \#$P$ of parts that are reachable by the human teammate only, on the size of the tree and the computation time (the time spent for planning and feasibility checks). For every tree, the number of nodes corresponding to each communication actions (i.e., Ask for help (K), Offer help (O), Confirm (C), and Request (R)) is reported as well.}
    \label{tab:cplanRH}
        \begin{minipage}{\textwidth}
		\begin{tabular}{ccccccccccccc}
			\hline\hline
			Inst. & \#$P$ & $L$ & $D=A+S+C$ & \multicolumn{4}{c}{Communication actions} & $DN$ & $BF$ & $N$ & \multicolumn{2}{c}{Time (sec)} \\
			&  &  &  & K & O & C & R &  &  &  & Plan  &  Checks \\ \hline
			6 & 2 & 21 & 8+7+13 & 2 & 2 & 7 & 2 & 133 & 4 & 367 & 657 & 41 \\
			7 & 3 & 23 & 9+6+15 & 3 & 1 & 7 & 4 & 128 & 5 & 590 & 1013 & 39 \\
			8 & 4 & 20 & 9+8+21 & 4 & 2 & 9 & 6 & 314 & 6 & 653 & 2095 & 53 \\
			9 & 5 & 24 & 8+9+20 & 5 & 2 & 8 & 5 & 467 & 7 & 989 & 4034 & 76 \\
			10 & 6 & 29 & 9+11+22 & 6 & 2 & 9 & 6 & 659 & 8 & 1534 & 6389 & 61 \\ \hline\hline
		\end{tabular}
        \end{minipage}\vspace{-0\baselineskip}
\end{table}

\begin{table}[t]
\caption{Experimental results demonstrating the effect of increasing the number \#$R$ of parts that are reachable by the robot only, on the size of the tree and the computation time (the time spent for planning and feasibility checks). For every tree, the number of nodes corresponding to each communication actions (i.e., Ask for help (K), Offer help (O), Confirm (C), and Request (R)) is reported as well.} 	
\label{tab:cplanRR}
        \begin{minipage}{\textwidth}
		\begin{tabular}{ccccccccccccc}
			\hline\hline
			Inst. & \#$R$ & $L$ & $D=A+S+C$ & \multicolumn{4}{c}{Communication actions} & $DN$ & $BF$ & $N$ & \multicolumn{2}{c}{Time (sec)} \\
			&  &  &  & K & O & C & R &  &  &  & Plan  &  Checks \\ \hline
			11 & 2 & 17 & 6+7+5 & 1 & 1 & 1 & 2 & 78 & 4 & 249 & 422 & 38 \\
			12 & 3 & 16 & 7+6+7 & 1 & 1 & 2 & 3 & 94 & 5 & 312 & 444 & 39 \\
			13 & 4 & 17 & 9+8+5 & 1 & 1 & 1 & 2 & 203 & 6 & 389 & 613 & 35\\
			14 & 5 & 21 & 11+7+9 & 1 & 1 & 3 & 4 & 353 & 7 & 411 & 965 & 42 \\
			15 & 6 & 18 & 13+7+5 & 1 & 1 & 1 & 2 & 399 & 8 & 509 & 1090 & 46 \\ \hline \hline
		\end{tabular}
        \end{minipage}\vspace{-.4\baselineskip}
\end{table}

\medskip  \noindent{\em Problem Instances.} We evaluate the results by computing hybrid conditional plans for 15 table assembly instances. We consider an initial table assembly setting with two unassembled table legs (one leg only accessible to the robot and the other only accessible to the human teammate), one unassembled foot (which is a dangerous object for the human, as it has a sharp screw nail and is accessible both to the human teammate and the robot) and a table top. In Table~\ref{tab:cplanEE}, we increase the number of dangerous objects on the table and examine how it will effect the tree size and the computation time. In Table~\ref{tab:cplanRH}, we increase the number of objects reachable to the human teammate,  while in Table~\ref{tab:cplanRR} the number of objects reachable by the robot is increased.

In Tables~\ref{tab:cplanEE}--\ref{tab:cplanRR}, the size of the tree is represented by the following parameters: the total number $L$ of leaves, the maximum length $D$ of a branch from the root to a leaf, and the number $A$ of actuation, $S$ of sensing and $C$ of communication actions in that branch, the total number $DN$ of decision nodes that denote sensing actions and nondeterministic communication actions, the maximum branching factor $BF$ (i.e., the maximum number of sensory outcomes), the total number $N$ of nodes in the tree (i.e., the size of the tree). We report the total computation time for the hybrid plan, as well as the time spent on the feasibility checks.

\medskip \noindent{\em Discussion of Results.} Several observations can be made from Tables~\ref{tab:cplanEE}--\ref{tab:cplanRR}.
\smallskip

The computation time of a hybrid conditional plan increases as its size increases. For instance in Table~\ref{tab:cplanEE}, a  hybrid conditional plan for Instance~1 (that consists of 104 decision nodes and 17 different hybrid sequential plans with a makespan up to 25) is computed in about 10 minutes, while a  hybrid conditional plan for Instance~5 (that consists of 511 decision nodes and 29 different hybrid sequential plans with a makespan up to 42) is computed in about 100~minutes. The increase in computation time is not surprising since, even for polynomially bounded plans with limited number of nondeterministic actions, the complexity of conditional planning is $\Sigma^P_2$-complete~\cite{BaralKT99}. On the other hand, note that the plan is computed offline considering all possible contingencies, and thus no time is spent for planning during execution.

The average computation time of a branch of the tree, which represents a possible hybrid sequential plan to reach the goal, is the total CPU time divided over $L$. For Instance~1 in Table~\ref{tab:cplanEE}, this time is around 3~minutes. This suggests that, if a hybrid sequential plan of actuation actions were computed instead of a hybrid conditional plan, then replanning would take around 3~minutes for Instance~1. Such (re)planning times are not acceptable  while communicating with a human. Therefore, computing an offline hybrid conditional plan that involves communications, in advance and by considering all possible contingencies, is advantageous  for collaborative tasks.

During hybrid conditional planning, the computational time spent for the feasibility checks is small compared to the planning time. In Table~\ref{tab:cplanEE}, hybrid conditional plan for Instance~1  is computed in about 10~minutes, while about 30~seconds is spent for feasibility checks. Similarly, hybrid conditional plan for Instance~5 is computed in about 100~minutes, while only 1~minute of this computation time is attributed to the feasibility checks.

In Table~\ref{tab:cplanEE}, as the number of unsafe objects ($\#U$) increases from 2 in Instance~1 to 6 in Instance~5, the number of nodes in the conditional plan and the computation time increase, from 246 to 1123 and from 588~sec to 5873~sec, respectively. Furthermore, the number of communication actions (O) to offer help to human increases significantly,  compared to the instances in Tables~\ref{tab:cplanRH} and~\ref{tab:cplanRR}.
This is expected, as the robot is required to offer help to improve safety of the operator.

In Table~\ref{tab:cplanRH}, as the number of objects reachable by the human teammate ($\#P$) increases from 2 in Instance~6 to 6 in Instance~10, the number of nodes in the conditional plan and the computation time increase, from 367 to 1534 and from 657~sec to 6389~sec, respectively. Furthermore, the number of communication actions (R) to request human help increases significantly,  compared to the instances in Tables~\ref{tab:cplanEE} and~\ref{tab:cplanRR}. Such an increase is expected, as the robot is required to request help for objects that are unreachable or close to the human.

In Table~\ref{tab:cplanRR}, as the objects which are reachable by the robot ($\#R$) increases from 2 in Instance~11 to 6 in Instance~15, the number of nodes in the conditional plan and the computation time increase, from 249 to 399 and from 422~sec to 1090~sec, respectively. Note that the increase in tree size and the computational time are significantly lower compared to Instances 1--10 in Tables~\ref{tab:cplanEE} and~\ref{tab:cplanRH}, as the number of all communication actions has been significantly reduced. The decrease in communication actions is expected since, in these instances, the robotic tasks can be performed without the need for communication with the human teammate.

\begin{figure}[b]
	\centering
	\resizebox{1\columnwidth}{!}{\includegraphics{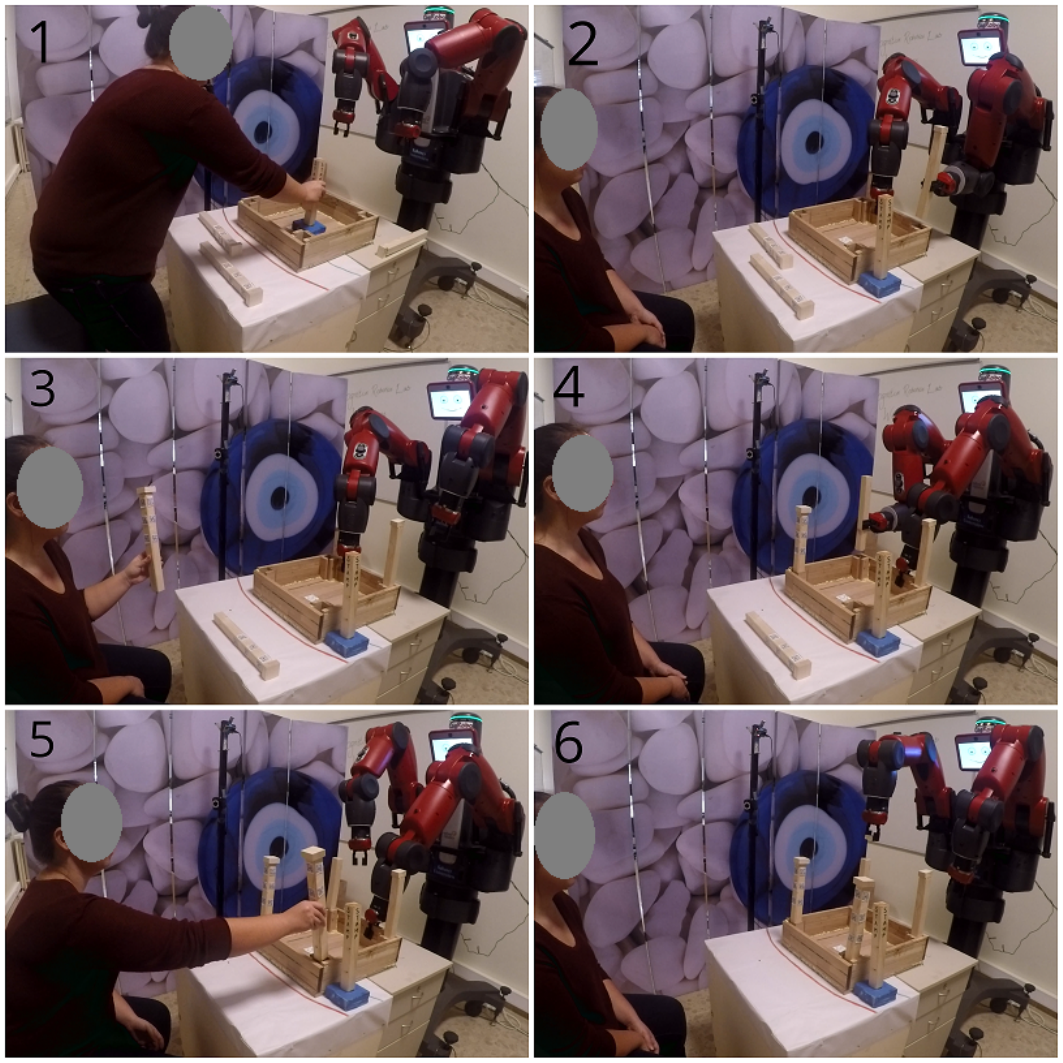}}
    \vspace{-1.25\baselineskip}
	\caption{Collaborative table assembly with a volunteer}
	\label{fig:snapshots} 
\end{figure}

\section{Collaborative Assembly Planning: A Real World Application}

Collaborative table assembly domain has aslo been tested through real-world applications with human volunteers.  During the physical executions, a volunteer and a bi-manual (Baxter) robot stand on the two opposite sides of a bench facing each other, as shown in the Figure~\ref{fig:snapshots}. The bench is divided into three regions: a shared region to which both the human teammate and the robot have access, a robot-only region and a human-only region.

In the collaborative table assembly task considered, a fully assembled table consists of a top, four equal length legs and four matching feet. Initially, human-robot team is presented with a set of legs of varying lengths (e.g., short, tall) and a set of feet of different shapes (i.e., square, triangle, circle) on different regions of the bench. A foot can be attached to a leg, if the shape of the foot matches with the hole in the leg. The robot has partial knowledge about the shapes of the feet and the connection types of the legs.

Since the task is collaborative, the robot is required to accommodate the uncertainties not only due the presence of a dynamic environment, but also due to the presence of human. The robot has to decide for a final configuration that precisely describes the desired product (i.e., which legs are assembled to the table top such that the table is stable, and which feet are connected to those legs), and to generate a plan of actions to reach the final configuration considering all the contingencies and human actions.

A fixed camera with a field of view of the whole scene was used to detect and track any changes and to monitor the execution of actions. All objects in the scene were labeled with QRcode markers to simplify object detection and tracking tasks. 3D object configurations were tracked online using an automated pattern-based object tracker.

During execution, the collaborative human teammate was instructed to stay within the field of view of the camera and to avoid occlusions to ensure that QRcode markers were visible by the camera at all times. Similar to all other moveable objects in the scene, QRcode markers were used to identify and track any parts held by the human teammate and to monitor the actions of the human.

The natural language communication was automated using Google Translate's text-to-speech API. In particular, for natural language communication, Python gTTS (Google Text-to-Speech) Library, which serves as a command-line interface tool to Google Translate’s text-to-speech API, was utilized. Furthermore, Google Speech API was used to recognize the responses of the human teammate.

An offline hybrid plan consists of not only a sequence of actions, but also collision-free paths that enable feasible execution of these actions. For real-time execution, these paths were provided to reference trajectory generation module of the Baxter robot in the order they are planned, such that the Baxter robot can follow trajectories along these collision-free paths under closed-loop motion control.

The planning, perception, control, and execution monitoring modules were integrated using the Robot Operating System (ROS). Figure~\ref{fig:snapshots} presents snapshots from physical execution with a volunteer.

In Snapshot~1, the robot explains that since the stamp is too close to the human and it is safer if she can stamp the table; in Snapshot~2 the robot executes an assembly task; in Snapshot~3 the robot senses that the human is holding a leg and confirms whether she wants to assemble it; in Snapshot~4 after the human completes her assembly, the robot assembles another leg; in Snapshot~5 robot asks human help to assemble a leg, as it is not feasible for the robot to reach the leg; in Snapshot~6 the robot picks a foot with the sharp nail (that is dangerous task for human) to assembles it to the leg. An annotated video of dynamic simulation of a sample collaborative assembly instance with Baxter robot is available at~\url{https://youtu.be/Bf6X8GLSamo}.


\section{Related Work}

\noindent{\em Collaborative Assembly Planning}
In typical assembly planning, no human-robot interaction is considered and uncertainties may exist only due to the incomplete knowledge of the world. However, human-robot collaboration is concerned with the uncertainty not only due to the incomplete knowledge about the state of the world but also due to the incomplete information about humans' actions, behavior, intentions, belief and  desires.

To reveal knowledge about the humans' mental state, communication is necessary. Human-robot communications have been used to guide collaborative planning, before the planning takes place, or after planning, that is, during the execution of the plan. For instance, in~\cite{kim2017collaborative}, communication between human and robot takes place before planning at a strategic level. While planning, they consider user's preferences to guide the planner. Experiments have been conducted in~\cite{unhelkar2014comparative} where human-robot communication takes place during the execution of fetch and deliver tasks. This study compares the performance of human while robot assistants help the worker, who is assembling a part, by fetching and delivering components. The work in~\cite{lasota2015} focuses on the motion level robot adaptation for safe close proximity human-robot collaborative assembly tasks.

Our approach is different from the above mentioned approaches, as we consider communication actions while planning for collaborative task. It is desirable to ensure task fluency, as we do not need to re-plan according to human behaviors and intentions since we plan for each possible communication contingency beforehand. It is also preferable because for each planned communication, we can provide evidence based explanations.

\noindent{\em Dialog Planning}
Human-robot interactions in natural language have been investigated by dialog-based approaches~\cite{petrick2013planning,giuliani2013comparing,tellex2014asking}. Some of these approaches use conditional planning~\cite{petrick2013planning}, some use branching plans~\cite{sebastiani2017dealing}, and some
use policy generation~\cite{grigore2016constructing} to incorporate communication actions in plans to obtain further knowledge. For instance, \citeN{petrick2013planning} and~\citeN{giuliani2013comparing} consider queries to learn what type of drink the human wants so that the robot prepares the customer's order  accordingly. In their approach, human does not perform any actions that can change the world state. \citeN{sebastiani2017dealing} consider queries to negotiate which tasks will be performed by the robot or the human. In this work, negotiation actions are not formalized as nondeterministic actions as part of the domain description, and thus the contingencies in communications are generated by an algorithm as execution variables. In~\cite{grigore2016constructing}, authors consider queries to reduce state estimation uncertainty in policy generation. Their goal is to assist the human rather than to plan for completion of a task collaboratively. Different from these related work, our goal is to plan for collaborative actions, and we consider a richer set of communication tasks. We formalize all the communication actions as part of the domain description, and utilize them as part of conditional planning.

Studies~\cite{petrick2013planning,giuliani2013comparing} are most related to our work, because communication actions are modeled formally as sensing actions and utilized while planning, for the purpose of constructing a dialogue: the robot communicates with human and serves them the requested drink. Our proposed approach utilizes communication for collaborative hybrid planning where human and robot perform actuation actions to reach a common goal and are aware of each other's intentions through observation and verbal communication. Collaborative tasks require richer communication actions, as observed above. Also, the representation language we use allows us to formalize commonsense knowledge.

The research work on Hierarchical Agent-based Task Planner (HATP) extended in~\cite{sebastiani2017dealing} to generate conditional plans for human-robot collaborations by adding on-line negotiations is also closely related to our approach. In this work, they generate shared plans including sensing actions for human-robot interactions and collaborative actions. Our method does not negotiate on-line at every step of the task by asking who is going to perform which task but computes an off-line hybrid conditional plan before execution.

In particular, we compute a hybrid conditional plan for actuation, sensing, and communication actions and perform those actions only when needed. For instance, while executing a task, if the robot senses that human pro-actively takes an initiative for a task, it confirms human intention, otherwise it continues performing its own task. If the robot is unable to perform a task (verified via a feasibility check), it can ask help from the human teammate. Human preferences may change from person to person: hence, due to this, we allow for specifying safety and verbosity level of plans to be generated.


\section{Conclusion}

We have introduced a novel method for collaborative assembly planning in uncertain and human-centric environments, using hybrid conditional planning based on ASP. This contribution is important for human-robot interactions from the following perspectives:

\begin{itemize}
\item Formal modeling of communication actions, embedded with formal representation of commonsense knowledge and low-level geometric checks, helps the robots to better understand when to communicate and how, as part of planning their actions. This is important for more effective collaborations of human-robot teams.
\item Offline planning of actions considering all contingencies with respect to outcomes of communication actions reduces the number of online replannings (as observed for sensing actions), and thus provides a more natural communication with human teammates.
\item Our formal modeling of actuation, sensing, and communication actions take safety concerns into account, utilizing hard and weak constraints of ASP. This is crucial for ensuring safety of human-robot collaborations.
    \item Our use of logic programming paradigm ASP for collaborative assembly planning provides a formal method for human-robot interaction studies. Investigating the use of such logic-based and provable methods is important for trustability of AI and robotic applications.
\end{itemize}

Our study also contributes to logic programming by extending its applications to another exciting, yet challenging area of robotics.

Based on the motivating empirical results and real-world applications on the furniture assembly domain,  our ongoing work includes extending the types of communication actions for more effective human-robot teams.

\bibliographystyle{acmtrans}

\label{lastpage}
\end{document}